\documentclass[10pt,journal,compsoc]{IEEEtran}
\usepackage{hyperref}
\usepackage[T1]{fontenc}
\usepackage[pdftex]{graphicx}
\usepackage{bm}
\usepackage{mathrsfs}
\usepackage{amsmath}
\usepackage{amssymb}
\usepackage{algorithm}
\usepackage{algorithmic}
\usepackage{array}
\usepackage{bm}
\usepackage{booktabs}
\usepackage{caption3}   
\usepackage{url}
\usepackage{multirow}
\usepackage{xcolor}
\usepackage[caption=false,font=footnotesize]{subfig}
\usepackage{pifont}
\usepackage{amsmath}
\usepackage[cmintegrals]{newtxmath}

\ifCLASSINFOpdf
\else
\fi
\ifCLASSOPTIONcompsoc
  \usepackage[nocompress]{cite}
\else
  \usepackage{cite}
\fi

\begin{document}
%
\title{Unsupervised Image Fusion Method based on Feature Mutual Mapping}
%
%
%
\author{Dongyu Rao,
        Xiao-Jun~Wu\textsuperscript{*},
        Tianyang~Xu,
        Guoyang Chen
\IEEEcompsocitemizethanks{\IEEEcompsocthanksitem D. Rao, X.-J. Wu (\textsl{Corresponding author}), and G. Chen are with the School of Artificial Intelligence and Computer Science, Jiangnan University, Wuxi, China. (e-mail: raodongyu@163.com, wu\_xiaojun@jiangnan.edu.cn).
\IEEEcompsocthanksitem {T. Xu is with the School of Artificial Intelligence and Computer Science, Jiangnan University, Wuxi 214122, P.R. China and the Centre for Vision, Speech and Signal Processing, University of Surrey, Guildford, GU2 7XH, UK. (e-mail: tianyang\_xu@163.com)}}
}

\markboth{Journal of \LaTeX\ Class Files.}%
{Shell \MakeLowercase{\textit{et al.}}: Bare Demo of IEEEtran.cls for Computer Society Journals}

\IEEEtitleabstractindextext{%
\begin{abstract}
Deep learning-based image fusion approaches have obtained wide attention in recent years, achieving promising performance in terms of visual perception.
However, the fusion module in the current deep learning-based methods suffers from two limitations, \textit{i.e.}, manually designed fusion function, and input-independent network learning. 
In this paper, we propose an unsupervised adaptive image fusion method to address the above issues.
We propose a feature mutual mapping fusion module and dual-branch multi-scale autoencoder. 
More specifically, we construct a global map to measure the connections of pixels between the input source images. 
Besides, we design a dual-branch multi-scale network through sampling transformation to extract discriminative image features. 
We further enrich feature representations of different scales through feature aggregation in the decoding process. 
Finally, we propose a modified loss function to train the network with efficient convergence property. 
Through sufficient training on infrared and visible image data sets, our method also shows excellent generalized performance in multi-focus and medical image fusion.
Our method achieves superior performance in both visual perception and objective evaluation. 
Experiments prove that the performance of our proposed method on a variety of image fusion tasks surpasses other state-of-the-art methods, proving the effectiveness and versatility of our approach.
\end{abstract}

}

\maketitle
\IEEEdisplaynontitleabstractindextext
\IEEEpeerreviewmaketitle
\IEEEraisesectionheading{\section{Introduction}}
\IEEEPARstart{I}{mage} fusion is an essential image processing technique that aims to fuse input images with different information into one output image containing more perceptive and salient visual clues.
The image fusion process requires the use of appropriate feature extractors and fusion strategies to achieve the ideal fusion effect \cite{li2017pixel}. 
With effective capacity in processing multi-source visual inputs, image fusion has a wide range of applications and excellent development prospects in photo processing, visual tracking \cite{li2020mdlatlrr, xu2019learning,xu2021adaptive,xu2020accelerated} and video surveillance, etc.

Image fusion currently includes multiple sub-tasks, such as infrared and visible image fusion \cite{jin2017survey}, multi-focus image fusion \cite{liu2017multi}, multi-exposure image fusion \cite{xu2020mef}, etc.. 
Fusion models for different tasks have a certain versatility as they share the same underlying function of integrating multiple images.
Therefore, image fusion algorithms can be simply divided into two types: traditional methods \cite{li2019discriminative,li2017multi,pajares2004wavelet,li2020laplacian,zhou2016perceptual,zheng2006nearest,chen2018new,luo2016novel,luo2017image,li2011no,wang2003initial,sun2011quantum,sun2019effective} and deep learning-based methods \cite{li2018densefuse,li2018infrared,ma2019fusiongan,zhao2018multi,fu2021image}.

The traditional methods are non-deep learning. 
According to the mathematics formulations, they are divided into the following categories:

(1) Multi-scale decomposition-based fusion: Multi-scale transformation is a widely used tool in image processing tasks. 
Infusion tasks, firstly, the source image is decomposed at multiple scales to obtain a multi-scale representation of the image. 
Then multi-scale representations from different images are fused according to specific fusion rules to obtain a fused multi-scale representation. 
Finally, the fused multi-scale representation is inversely transformed to obtain the fused image. 
Many algorithms that use multi-scale transforms have obtained promising fusion results, such as pyramid transform \cite{mertens2007exposure}, wavelet transform \cite{zhang1999categorization} and contourlet transform \cite{zhang2016adaptive}, etc..

(2) Sparse representation-based fusion: Sparse representation (SR) is an image representation technique, which has been applied in many image-related tasks. 
It selects a sparse linear combination of atoms from the over-complete dictionary learned from high-quality images to represent the input. 
In the fusion task, the input image is divided into patches and then decomposed to obtain the corresponding sparse coefficients \cite{chen2014image,nejati2015multi,kim2016joint}. 
After decomposition, the fusion rules are applied to the coefficients of different source images to obtain the fused sparse coefficients. 
Finally, the fusion image is reconstructed through the fusion coefficient and dictionary.

(3) Fusion in other domains: In addition to methods based on multi-scale decomposition and sparse representation, there are other methods such as low-rank representation (LRR), subspace, etc. 
In particular, LRR can extract the features of the image in a specific domain (low-rank) for fusion to obtain a fused image \cite{li2017multi}. 
The method based on subspace is to map the image from high-dimensional to low-dimensional, eliminating the interference of redundant information and facilitating the acquisition of the internal structure of the image \cite{bavirisetti2017multi,mou2013image}.

Although traditional methods can achieve acceptable fusion effects, the limitations of traditional methods for feature representation result in the focus on the field of deep learning\cite{chen2017deep,zhang2016deep, wu2015weakly}.
In 2017, Liu \cite{liu2017multi} et al. proposed a convolutional neural network method that requires training for multi-focus image fusion.
In 2018, Li \cite{li2018infrared,li2019infrared} et al. used a pre-trained model to extract image deep features for infrared and visible images fusion tasks.  
These methods show superior performance in terms of accuracy and speed than traditional methods.
This also proves the feasibility of the image fusion method based on deep learning techniques.
With the deepening of research based on deep learning, many image fusion methods based on deep learning have been proposed, achieving improved fusion results \cite{fu2021dual}.
In 2018, Li \cite{li2018densefuse} et al. proposed a method to fuse infrared and visible images based on an auto-encoder. 
Guo \cite{guo2018fully} et al. proposed a multi-focus image fusion method using a fully convolutional neural network.
Ma \cite{ma2019fusiongan} proposed a method that uses a generative adversarial network (GAN) to achieve infrared and visible image fusion for the first time. 
Li \cite{li2021rfn} et al. proposed an infrared and visible image fusion method using adaptive fusion rules.

However, image fusion methods based on deep learning still suffer from the following issues: (1) Due to the limitation of the network structure, the extracted information is not comprehensive enough to support high-level salience. (2) The design of the fusion strategy is rough thus cannot provide fine-grained multi-source interaction. 
(3) High-quality fusion results cannot always be guaranteed, challenged by image blur and loss of details.

In order to overcome the above shortcomings, we propose an unsupervised image fusion method based on feature mutual mapping in this paper. 
In order to better extract the features of the image, we use multiple res-blocks \cite{he2016deep} to build the encoder, and obtain the multi-scale information of the image through down-sampling. 
In the fusion stage, we are inspired by the self-attention mechanism and propose feature mutual mapping for adaptive feature fusion. 
In the decoding stage, we enrich the information of the fused features through feature aggregation \cite{yu2018deep}, generating the fused image.
In addition, our model and loss function can be directly generalized to other image fusion tasks, such as multi-focus image fusion and medical image fusion.
Experiments show that our fusion method exhibits excellent performance on a variety of fusion tasks.

The main innovations of the proposed method include:
\begin{itemize}
\item We design an auto-encoder network that aggregates multi-scale representations to enhance the output features. 
\item We propose a feature mutual mapping module to adaptively learn the fusion rules, formulating a unified framework for a variety of fusion tasks.
\item A single loss function is used during training, generating excellent fusion performance, and accelerating the training convergence.
\end{itemize}

The paper is organized as follows:
In Section~\ref{relatedwork}, we introduce the studies related to the proposed fusion method in this paper.
In Section~\ref{method}, we present the framework and training details of the proposed method. 
The experimental results and their analysis are reported in Section~\ref{experiment}.
Finally, the conclusions are drawn in Section~\ref{conclusion}.

\section{Related Works}
\label{relatedwork}
In recent years, many image fusion methods based on deep learning have been proposed\cite{liu2018deep}. 
Researchers have made different attempts to modify the network structure, fusion module, and loss function to obtain a fused image with better performance.
In this section, firstly, we review several image fusion methods based on deep learning in recent years. 
Then we introduce the application of the self-attention mechanism in the computer vision field.
\subsection{Deep Learning-based Image Fusion Methods}
In 2017, Liu et al.\cite{2018Infrared} propose an image fusion method based on a convolutional neural network. 
In this paper, infrared and visible images are used as source images, generating the output containing visible details and infrared features. 
Firstly, a convolutional network is used to obtain a weight map that integrates information from two source images.
According to the different characteristics of the infrared image and visible light image, multi-scale fusion is carried out through the image pyramid.          
Finally, the fusion image is obtained through Laplacian pyramid reconstruction.
Experimental results show that the method achieves state-of-the-art results. 
However, this method only obtains the weight map through the convolutional neural network, without performing deep analysis.

In 2018, Ren et al. \cite{ren2018infrared} use the VGG network to extract the features of infrared and visible source images, accompanied by a corresponding loss function to complete the fusion task. 
The fusion image is composed of general features extracted by the convolutional neural network in different proportions.
This method has achieved good results in both subjective and objective evaluation indicators. However, due to the limitations of training methods and loss functions, this method is only suitable for infrared and visible image fusion tasks.

Li et al. \cite{li2018densefuse} propose an autoencoder structure based on deep learning, applying different fusion strategies to the encoded image features.
This method uses public data sets to train encoders and decoders, alleviating the limitation of insufficient training data for infrared and visible image fusion tasks. 
Since the fusion strategy does not participate in training, the autoencoder can also be used as a general framework for different tasks with different fusion strategies. 
However, the fusion strategy is not synchronized with the training, which means that the method requires manual selection for each fusion task.

\begin{figure*}[htbp]
	\begin{center}
		\includegraphics[width=0.9\linewidth]{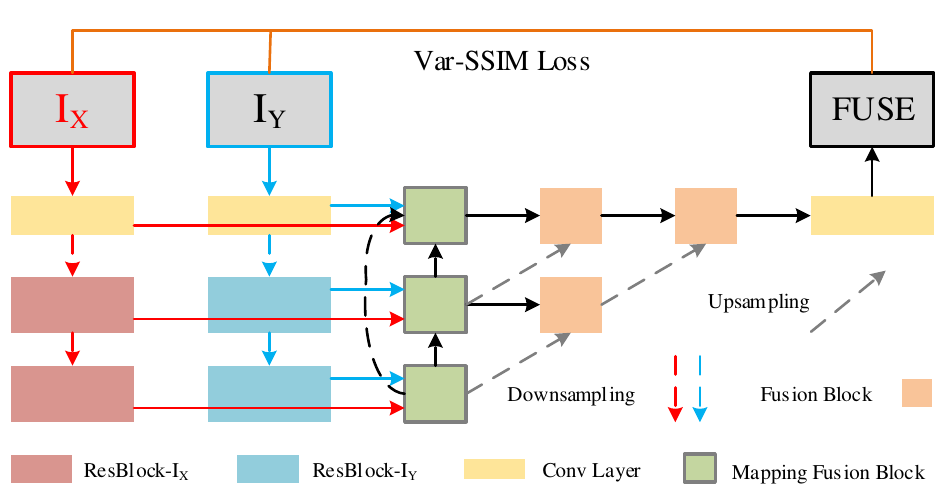}
		\caption{The framework of our method.}\label{main-fram}
	\end{center}
\end{figure*}

In 2019, Ma et al. \cite{ma2019fusiongan} introduce the method of the generative adversarial network into the image fusion task. 
This method uses a generator to generate a fusion image that retains the main infrared information and visible gradients. 
The discriminator is used to force the fused image to exhibit additional visible details.
Specifically, the fusion image generated by GAN is clean and clear which is not sensitive to noise in the infrared image.
\subsection{Self-attention Mechanism}
In 2017, Lin et al. \cite{lin2017structured} propose the self-attention mechanism and apply it to the field of natural language processing. 
Subsequently, the work of Vaswani et al. \cite{vaswani2017attention} extends the self-attention approach with widespread attention and application. 
Similar to the attention mechanism, the purpose of the self-attention mechanism is to integrate comprehensive information from a large number of connected nodes.
The difference is that the self-attention mechanism reduces the dependence on external information, focusing on capturing the internal correlation of features.
Its application in language processing is mainly to solve the problem of long-distance dependence by calculating the mutual influence between words.

In the field of computer vision, self-attention can effectively capture the long-range dependency between different positions \cite{wang2018non}. 
Zhang et al. \cite{zhang2019self} use self-attention for image generation tasks and propose SAGAN. 
SAGAN adds a self-attention module on the basis of SNGAN \cite{miyato2018spectral}, so that the generation of each position is no longer solely dependent on other positions adjacent to this position, while introducing a position that is farther and has a greater amount of information.
\section{The Proposed Fusion Method}
\label{method}
In this section, we introduce our method in detail.
Firstly, the overall framework of our method is presented in section \ref{sec3-1}. 
Then, we introduce the fusion method based on the image mapping relationship of different modalities, which is derived from the self-attention mechanism in section \ref{sec3-2}.
Finally, the detail of the training phase is shown in section \ref{sec3-3}.
\subsection{Framework of Network}
\label{sec3-1}
As shown in Fig. \ref{main-fram}, our network consists of three parts, including encoder, fusion module, and decoder. 
The red and blue rectangles on the left of Fig. \ref{main-fram} represent the encoders for different source images respectively.
The green square represents the mapping fusion module. The other shapes denote the decoder.

In Fig. \ref{main-fram}, “I$_X$” and “I$_Y$” represent different source images respectively.
Unlike other methods \cite{li2020nestfuse}, our network does not share the network weights of the two modalities, as images of different modalities present different information distribution.
Therefore, from the perspective of feature mapping, two different feature spaces for each modality can better retain the unique information of a modal than sharing the same feature space.
In order to characterize the corresponding modal information through the trained neural network, we set up two symmetrical encoder branches for infrared and visible images. 
Each encoder branch includes a 3$times$3 convolution and two res-blocks. 
In order to obtain a feature map of different scales, down-sampling is performed between each block to change the size of the feature.

\begin{figure}[h]
	\begin{center}
		\includegraphics[width=1\linewidth]{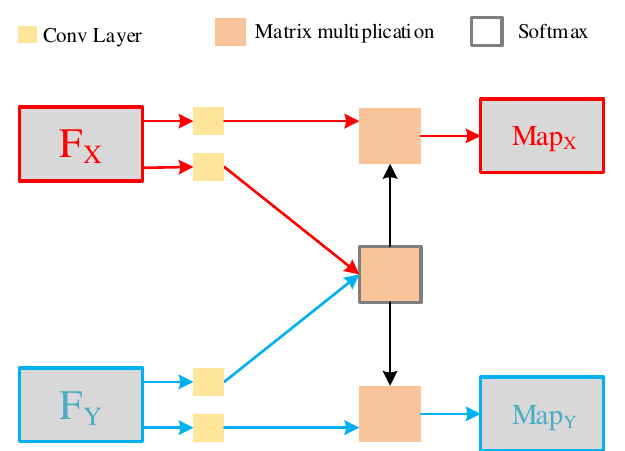}
		\caption{The framework of mapping fusion module.}\label{map-frame}
	\end{center}
\end{figure}

As the Fig.\ref{map-frame} shows, the mapping fusion block is an adaptive fusion method to find the feature mapping between source images. 
The input of the mapping fusion module is the extracted feature of two modal images. 
This module achieves a better fusion effect by finding the global relationship of each spatial position between one modal image and another modal image. 
The specific implementation of the mapping fusion module will be described in section \ref{sec3-2}.

For the decoder, each block consists of a concatenation and convolution operation. 
In the decoding process, we aggregate features of different scales through feature aggregation \cite{yu2018deep}, so that the fused image retains more multi-layer clues. 
Through a layer of convolution, the fused features are restored to the original shape and dimension, thus obtaining the final output.
\subsection{The Mapping Fusion Module}
\label{sec3-2}
In existing deep learning-based image fusion methods, commonly used fusion strategies include direct concatenation and manual fusion rule design, etc. 
Direct concatenation of cross-modal features can obtain better fusion results, while delivering information redundancy.
Manually designed fusion strategies are more flexible but difficult to find the optimal solution.
Therefore, we try to find a strategy that can adaptively realize the multi-modal fusion according to the relationship between the cross-modal information in the image fusion task. 

According to the analysis, we can conclude that performing direct concatenation fusion can not achieve fine-grained image information integration.
Besides, the fusion process can only capture the local information constrained in the receptive field that is limited by the size of the multi-layer convolution kernels. 

Therefore, we argue that finding the global relationship between different source image features is essential for generating adaptive fusion weights.
The characteristics of the self-attention mechanism are employed to implement the global relationship construction.
In the field of computer vision, self-attention usually determines the importance of the information in the image by calculating the relationship between each spatial point and other points on a single image \cite{bello2019attention}. 
In order to find the mapping relationship between the two modalities, we refer to the implementation of self-attention to involve the features of different modalities, performing a matrix multiplication operation to measure their degree of correlation. 
Then the result obtained by the dot product is input to a softmax operation and performs matrix multiplication with two modes separately.
Through the above operations, we have obtained the global positional relationship on one modal with respect to another modal feature.
We use the output map as the attention weight of the modal.
As shown in Eq. \ref{eq-fuse}, $Map_{X}$ and $Map_{Y}$ mean the map of different source images respectively.
$f_x$ and $f_y$ represent the features of different source images  respectively. 
$fuse$ is the feature obtained by fusion.
\begin{equation}\label{eq-fuse}
	fuse = Map_{X}*f_x + Map_{Y}*f_y. 
\end{equation}

The matrix multiplication operation requires a lot of computing resources, while the mapping relationship between the features can also well represent the mapping relationship between the two modal images, so we use the mapping fusion block for the image after two downsamplings.
For the shallow features, the deep  $Map_{X}$ and $Map_{Y}$ are upsampled, and then the shallow map is obtained through dense connection \cite{huang2017densely}.
For the shallow fusion module, the fusion feature is obtained by implementing the map and the corresponding feature according to Eq. \ref{eq-fuse}.
\subsection{Loss function}\label{sec3-3}
In the training phase, our method adopts an end-to-end training method. 
Through the learning of neural networks, we can get a multi-branch encoder for infrared visible images, a mapping fusion module, and the corresponding decoder. 
The structure of the network during training is shown in Fig \ref{main-fram}.

We propose an approach that relies on a single loss function to achieve the fusion task. 
Unlike the previous method \cite{li2018densefuse, hou2020vif} that uses multiple loss functions to jointly supervise the training stage, we can achieve a better fusion effect using only a single loss function. 
In addition, the simplification of the loss function also speeds up the training process.

The SSIM \cite{wang2004image} is a measure of the structural similarity between two images.
The calculation process of structural similarity is shown in Eq.\ref{eq-ssim}.
X, Y represent two images respectively. 
$\mu$ and $\sigma$ stand for mean and standard deviation respectively. 
$\sigma_{XY}$ means the covariance between X and Y. 
$C1$ and $C2$ are stability coefficients. 
\begin{equation}\label{eq-ssim}
	SSIM(X, Y) = \frac{(2\mu_X\mu_Y + C_1)(2\sigma_{XY} + C_2)}{({\mu_X}^2 + {\mu_Y}^2 + C_1)({\sigma_X}^2 + {\sigma_Y}^2 + C_2)}
\end{equation}

Due to the effectiveness of SSIM, it is used as the cost function in a variety of image fusion tasks \cite{ma2017multi,li2020nestfuse}. 
Since the structural similarity calculates the overall information of the image, neglecting the local visual connections. 
Therefore, simply using SSIM as the loss function cannot achieve a visually acceptable fusion effect.
Taking infrared and visible image fusion as an example, since there are more noises in the infrared images, we usually hope to get more details from the visible image, such that other losses and parameters are introduced to enhance this information \cite{li2020nestfuse}. 
Inspired by \cite{hou2020vif,yan2018unsupervised}, we combined SSIM with the variance of the image to obtain a modified fusion loss function. 

\begin{figure*}[t]
	\begin{center}
		\includegraphics[width=1\linewidth]{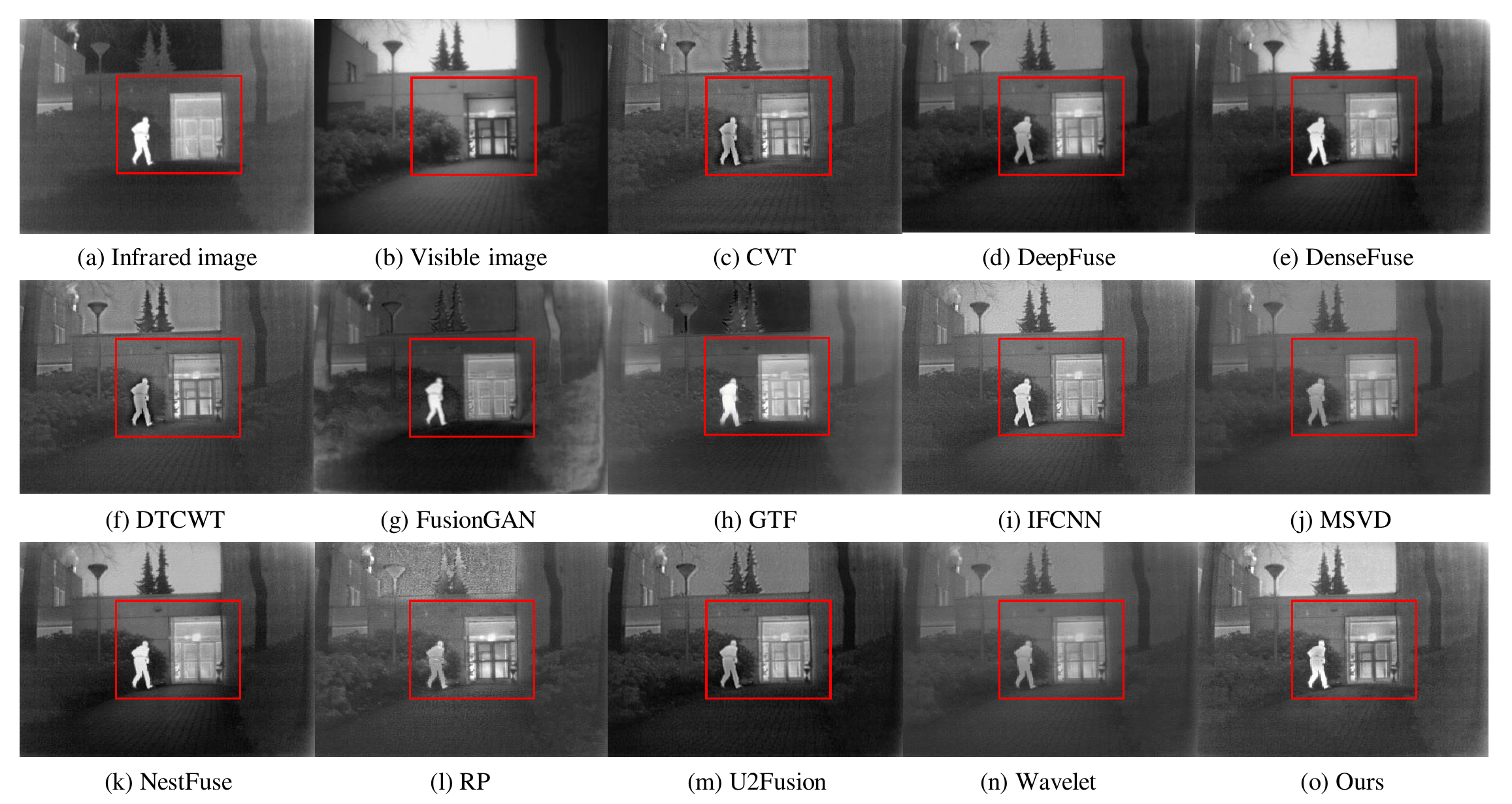}
		\caption{Infrared and visible image fusion experiment on “human” images.}\label{fig-ir1}
	\end{center}
\end{figure*}
\begin{figure}[htbp]
	\begin{center}
		\includegraphics[width=1\linewidth]{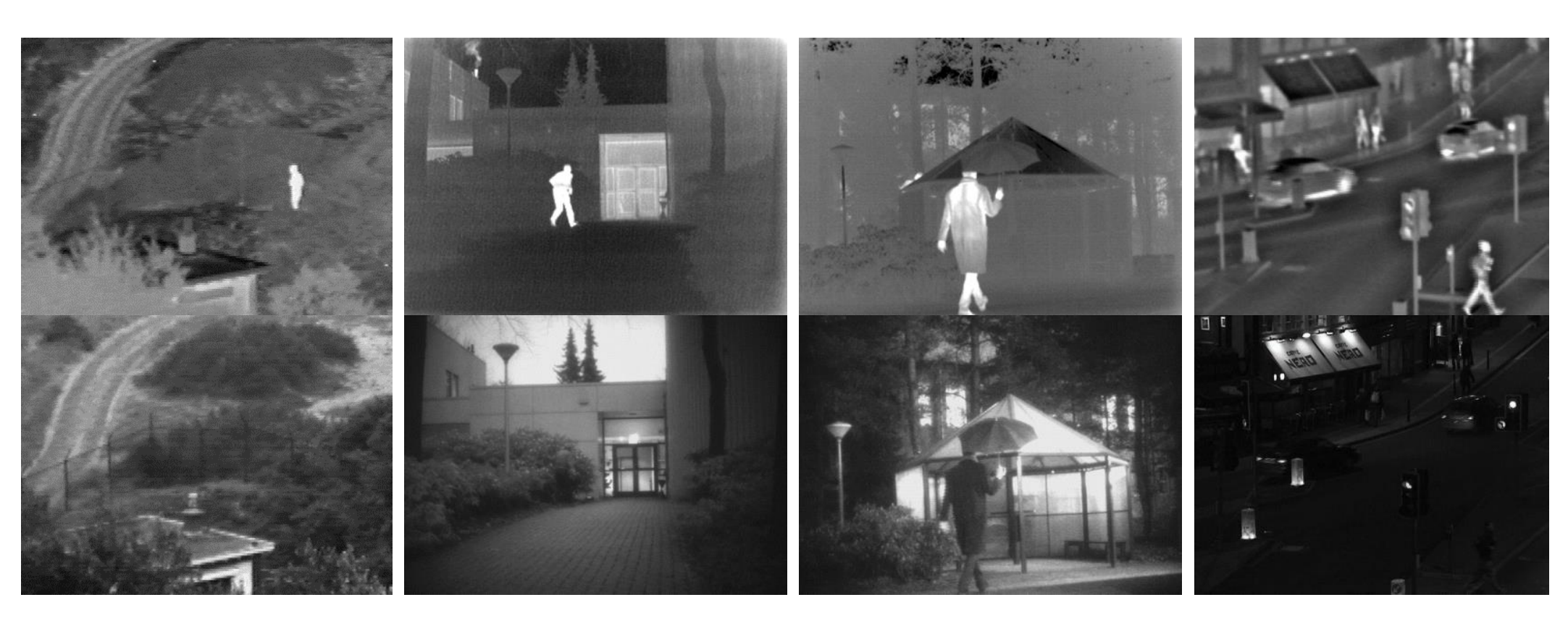}
		\caption{Four pairs of source images. The top row contains infrared images, and the second row contains visible images.}\label{fig-ir}
	\end{center}
\end{figure}

The variance of an image represents the distribution of data pixels in an image.
As shown in Eq.\ref{eq-var}, M and N represent the image size in the horizontal and vertical directions respectively. 
$\mu$ is the mean of the image. 
\begin{equation}\label{eq-var}
\sigma^2(X) = \frac{\mathop\Sigma\limits_{i=0}\limits^{M-1}\mathop\Sigma\limits_{j=0}\limits^{N-1}[X(i, j) - \mu]^2}{MN}
\end{equation}

$Var_-SSIM$ calculates the structural similarity between fused image and source image patches. $\sigma^2$ is the basis for choosing which source image to compare. As shown in Eq. \ref{eq-varssim}, $I_X$ and $I_Y$ represent two input images respectively. 
$I_F$ is a fused image. 
W is the sliding window, and its size is set to 11$\times$11.
Through the sliding window, the image is divided into multiple overlapping patches. 
We calculate the variance and the structural similarity of each patch. 
Finally, the patch with larger variance is selected as the source image, which structural similarity is also used to calculate the final structural similarity loss. 
The loss function is shown in Eq. \ref{eq-loss}.
\begin{equation}\label{eq-varssim}
Var_-SSIM(I_X, I_Y, I_F|W) = 
\left\{
\begin{array}{lr}
SSIM(I_X, I_F), \\
if \sigma^2(X) > \sigma^2(Y) &  \\
SSIM(I_Y, I_F), \\
if \sigma^2(Y) >= \sigma^2(X) &  
\end{array}
\right.
\end{equation}
\begin{equation}\label{eq-loss}
L_{var_-SSIM} = 1 - \frac{1}{N}\mathop\Sigma\limits^N\limits_{W=1} Var_-SSIM(I_X, I_Y, I_F|W)
\end{equation}
\section{Experiments and analyses}\label{experiment}
In this section, we verify our approach in different image fusion tasks.
First, we show the details of training and testing implementations in Sec \ref{train-test}.
Then, different evaluation criteria for image fusion tasks are shown, including subjective and objective evaluations in Sec \ref{eva}.
In Sec \ref{ir}, we show the performance of our method on the fusion of infrared and visible images.
In Sec \ref{focus} and Sec \ref{medical}, the performance of our method on the fusion of multi-focus images and medical images are reported respectively.
\subsection{Training and test details}\label{train-test}
In the training stage, we select 40,000 pairs of corresponding infrared and visible images from the KAIST \cite{hwang2015multispectral} data set as training data. 
The training image size is set to 256$times$256 pixels. 
During training, we use Adam as the optimizer and the learning rate is set to a constant 0.0001. 
The whole training process includes 4 epochs.
Our model is implemented with NVIDIA TITAN Xp and Pytorch. 

\begin{figure*}[t]
	\begin{center}
		\includegraphics[width=1\linewidth]{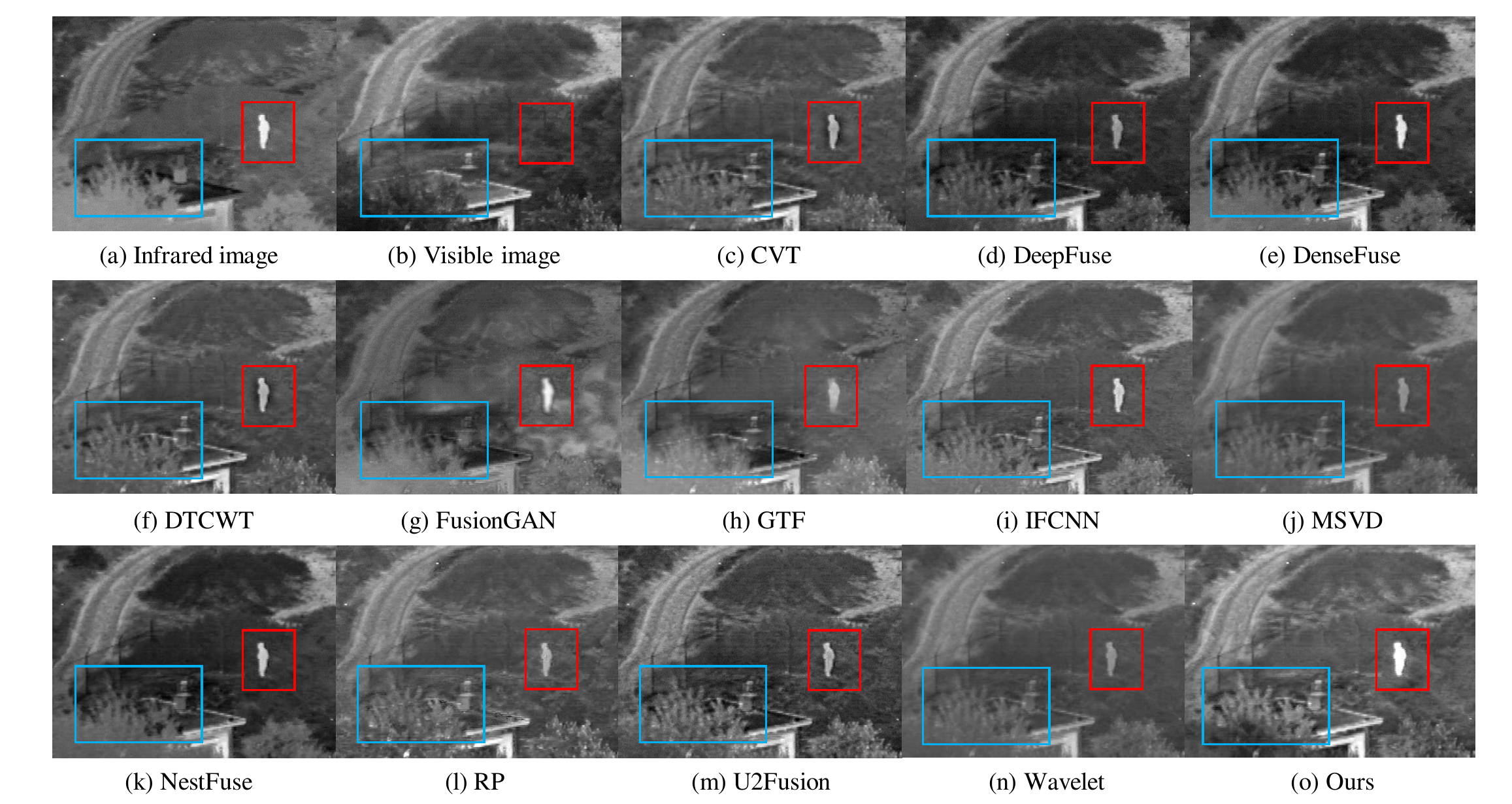}
		\caption{Infrared and visible image fusion experiment on “human and tree” images.}\label{fig-ir2}
	\end{center}
\end{figure*}
\begin{table*}[!t]
	\footnotesize
	\renewcommand{\arraystretch}{1.3}
	\caption{Quantitative evaluation results of infrared and visible image fusion tasks. The best two results are highlighted in {\textbf{bold}} and {\emph{italic}} fonts.}
	\label{eva-ir}
	\centering
	\begin{tabular}{ccccccccc}
		\hline
		Methods       & EI      & CE       & SF       & EN       & Q$^{ab/f}$     & ~MS$_-$SSIM & SD       & VIF       \\ 
		\hline
		RP            & 44.9054 & 1.3420   & \textbf{12.7249}  & 6.5397   & 0.4341   & 0.8404    & 63.2427  & 0.6420    \\
		Wavelet       & 24.6654 & 1.4278   & 6.2567   & 6.2454   & 0.3214   & 0.8598    & 52.2292  & 0.2921    \\
		DTCWT         & 42.4889 & 1.6235   & 11.1296  & 6.4791   & \emph{0.5258}   & 0.9053    & 60.1138  & 0.5986    \\
		CVT           & 42.9631 & 1.5894   & 11.1129  & 6.4989   & 0.4936   & 0.8963    & 60.4005  & 0.5930    \\
		MSVD          & 27.6098 & 1.4202   & 8.5538   & 6.2807   & 0.3328   & 0.8652    & 52.9853  & 0.3031    \\
		GTF           & 35.0073 & \textbf{1.1440}   & 9.5022   & 6.5781   & 0.4400   & 0.8169    & 66.0773  & 0.4071    \\
		DenseFuse & 36.4838 & 1.4015   & 9.3238   & \textbf{6.8526}   & 0.4735   & 0.8692    & \emph{81.7283}  & 0.6875    \\
		DeepFuse      & 33.8768 & 1.7779   & 8.3500   & 6.6102   & 0.3847   & 0.9138    & 66.8872  & 0.5752    \\
		IFCNN         & 44.9725 & 1.4413   & \emph{11.8590}  & 6.6454   & 0.4962   & 0.9129    & 73.7053  & 0.6090    \\
		FusionGAN     & 32.5997 & 1.9353   & 8.0476   & 6.5409   & 0.2682   & 0.6135    & 61.6339  & 0.4928    \\
		NestFuse & 37.5627 & 1.4463   & 9.5383   & \emph{6.8492}   & 0.4979   & 0.8816    & 80.2687  & 0.7018    \\
		U2Fusion      & \textbf{48.4915} & 1.3255   & 11.0368  & 6.7227   & 0.3934   & \emph{0.9147}    & 66.5035  & \emph{0.7680}    \\
		Ours          & \emph{45.5980} & \emph{1.2941} & 11.5680 & 6.8435 & \textbf{0.5581} & \textbf{0.9164}  & \textbf{93.0976} & \textbf{0.7897}  \\
		\hline
	\end{tabular}
\end{table*}

Through sufficient training on infrared and visible image data sets, we believe that the model has been sufficiently learning the feature extractor and fusion rules. 
Due to the similar characteristics of infrared and visible light images, multi-focus images, and medical image fusion tasks, we did not perform additional training on multi-focus image fusion and medical image fusion tasks. 
In the test, we used 10 pairs of images in \cite{li2018densefuse} as the test set of infrared and visible image tasks. 
Multi-focus and medical images in \cite{zhang2020ifcnn} are used as test sets for multi-focus image fusion and medical image fusion tasks, respectively.

\subsection{Subjective and Objective Evaluation}\label{eva}
When evaluating the fusion results, we use two evaluation criteria: subjective evaluation and objective evaluation.

Subjective evaluation starts from human vision to judge whether the fusion result conforms to visual habits, such as clarity, saliency information, etc.
Therefore, the subjective evaluation method is to put the results of different methods together and directly display them.

In terms of objective evaluation, we provide different evaluation methods for image quality, information, edge, etc. These are: Edge Intensity (EI) \cite{xydeas2000objective},  Cross Entropy (CE) \cite{kumar2013multifocus}, Spatial Frequency (SF) \cite{eskicioglu1995image}, Entropy (EN) \cite{roberts2008assessment}, quality of images ($Q^{ab/f}$) \cite{xydeas2000objective}, multiscale SSIM (MS$_-$SSIM), Standard Deviation of Image (SD) \cite{rao1997fibre}, Visual Information Fidelity (VIF) \cite{sheikh2006image}.

\begin{figure*}
	\begin{center}
		\includegraphics[width=1\linewidth]{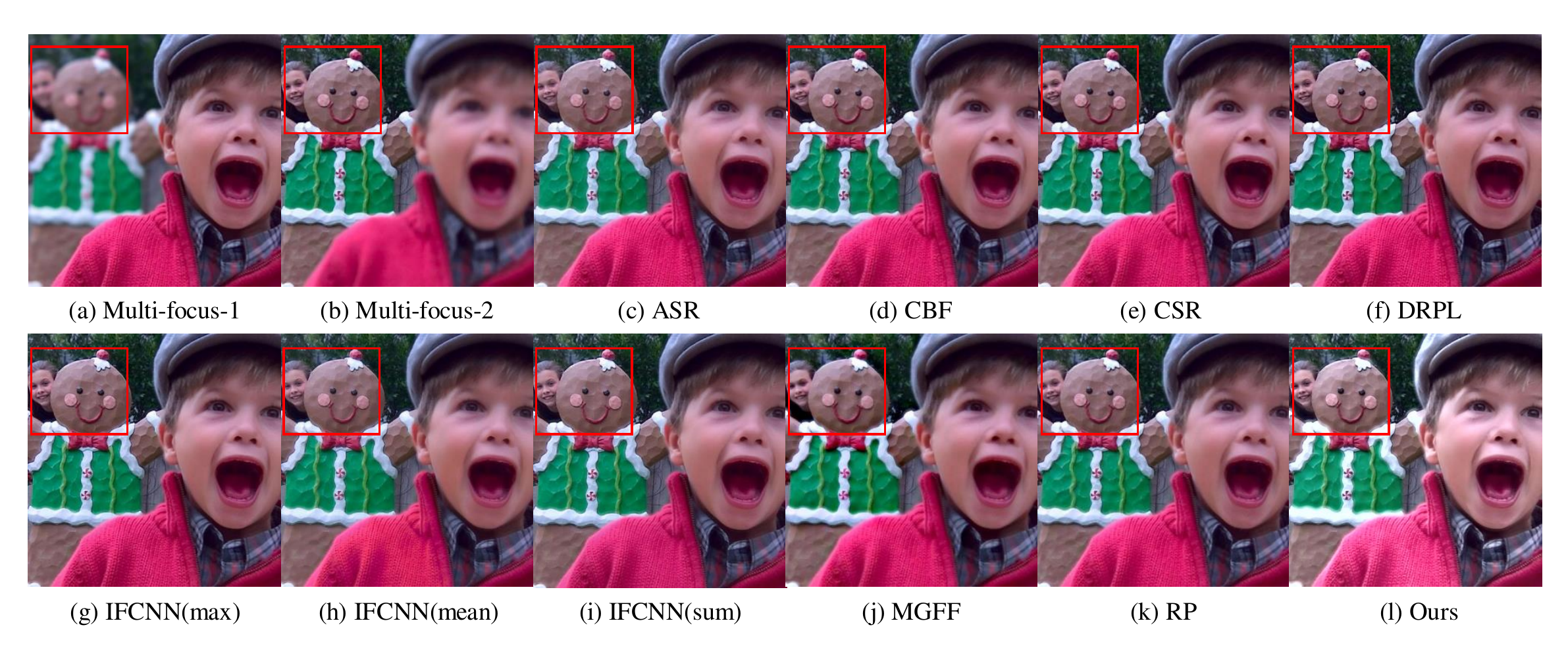}
		\caption{The multi-focus image fusion experiment results of “girl and toy” images.}\label{fig-focus1}
	\end{center}
\end{figure*}
\begin{table*}[!t]
	\footnotesize
	\renewcommand{\arraystretch}{1.3}
	\caption{Quantitative evaluation results of multi-focus image fusion tasks. The best three results are highlighted in {\color{red}{red}}, {\color{blue}{blue}} and {\color{brown}{brown}} fonts.}
	\label{eva-focus}
	\centering
	\begin{tabular}{cccccccc}
		\hline
		Methods     & EI      & SF      & EN     & SCD    & MI      & SD       & VIF     \\
		\hline
		ASR         & 69.0392 & 18.9739 & 7.5281 & 0.4839 & 15.0561 & 115.8590 & 1.0422  \\
		CBF         & 68.5800 & 18.5826 & 7.5278 & 0.4771 & 15.0555 & 115.8917 & 1.0668  \\
		CSR         & 69.8621 & 19.2245 & 7.5313 & 0.5292 & 15.0625 & 117.1057 & 1.1273  \\
		DRPL        & \color{blue}{\textbf{70.9735}} & \color{brown}{\textbf{19.4398}} & 7.5334 & 0.5446 & 15.0667 & 117.3645 & 1.1389  \\
		RP\_SP      & 70.7294 & \color{blue}{\textbf{19.4944}} & \color{blue}{\textbf{7.5389}} & 0.5645 & \color{blue}{\textbf{15.0778}} & 117.4334 & 1.1407  \\
		MGFF        & 63.1033 & 17.3024 & \color{red}{\textbf{7.5685}} & \color{brown}{\textbf{0.7480}} & \color{red}{\textbf{15.1370}} & \color{brown}{\textbf{120.3308}} & \color{brown}{\textbf{1.2356}}  \\
		IFCNN(mean) & 65.5623 & 17.7762 & 7.5120 & 0.5253 & 15.0241 & 115.3850 & 1.0541  \\
		IFCNN(max)  & \color{brown}{\textbf{70.7663}} & 19.4004 & 7.5319 & 0.5817 & 15.0639 & 117.3573 & 1.1339  \\
		IFCNN(sum)  & 69.2496 & 18.9037 & 7.5334 & 0.5649 & 15.0667 & 116.6219 & 1.0979  \\
		U2Fusion    & 69.2651 & 17.2710 & 7.5290 & \color{blue}{\textbf{0.9874}} & 15.0581 & \color{blue}{\textbf{123.8306}} & \color{blue}{\textbf{1.2799}}  \\
		Ours        & \color{red}{\textbf{77.1693}} & \color{red}{\textbf{21.3001}} & \color{brown}{\textbf{7.5381}} & \color{red}{\textbf{1.1038}} & \color{brown}{\textbf{15.0763}} & \color{red}{\textbf{135.1799}} & \color{red}{\textbf{1.4193}}  \\
		\hline
	\end{tabular}
\end{table*}
\begin{figure}[!t]
	\begin{center}
		\includegraphics[width=1\linewidth]{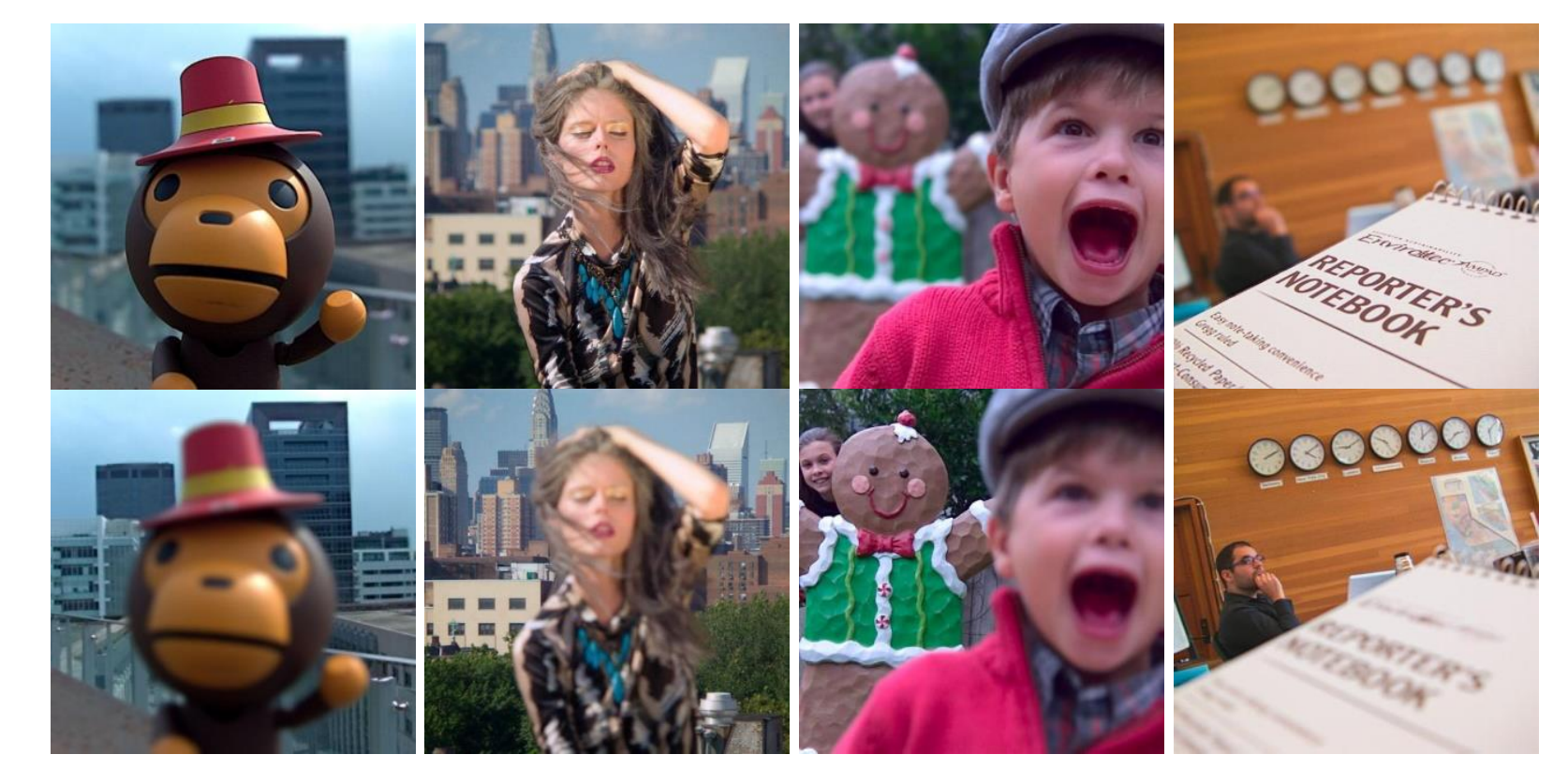}
		\caption{Four pairs of source images. The top row contains focus on the close-up images, and the second row contains focus on the distant images.}\label{fig-focus}
	\end{center}
\end{figure}

\subsection{Visible and Infrared Image Fusion}\label{ir}
In Fig.\ref{fig-ir}, we show some images in the infrared and visible image fusion task. 
The upper row is the infrared image, the lower row is the corresponding visible image.
The goal of infrared and visible image fusion is to combine the saliency information in the infrared image (the part of the highlighted object) with the detailed information of the visible image to obtain a fused image with better visual effects.
In the testing phase, we selected 10 pairs of infrared and visible images to be fused and evaluated by objective methods.

In Fig. \ref{fig-ir1}, we show a comparison of the visualization results of the involved 13 methods.
In the red frame, the infrared source image has a bright human body, and the visible image has the details of the door frame.
Other methods can only retain one modality of the contents, and the other modality is weakened by sufficient information fusion. 
The fused image obtained by our method can retain the highlights of the human body and the details of the door frame at the same time. 

In Fig. \ref{fig-ir2}, the bright human body in the infrared source image (red box) and the edge of the leaf in the visible image (blue box) are the fused objects. 
In other methods, the fusion of infrared and visible images by leaves is insufficient. 
At the same time, the saliency information in the fused image is not obvious, because the human body lacks visual clues in the corresponding area of the visible image. 
In our method, the feature map fusion module has learned the global relationship corresponding to the two source images. 
Accordingly, the red frame part retains the human body information highlighted in the infrared image, and the blue frame part blends the edges of the leaves and the subject in the two source images well.

\begin{figure*}[!t]
	\begin{center}
		\includegraphics[width=1\linewidth]{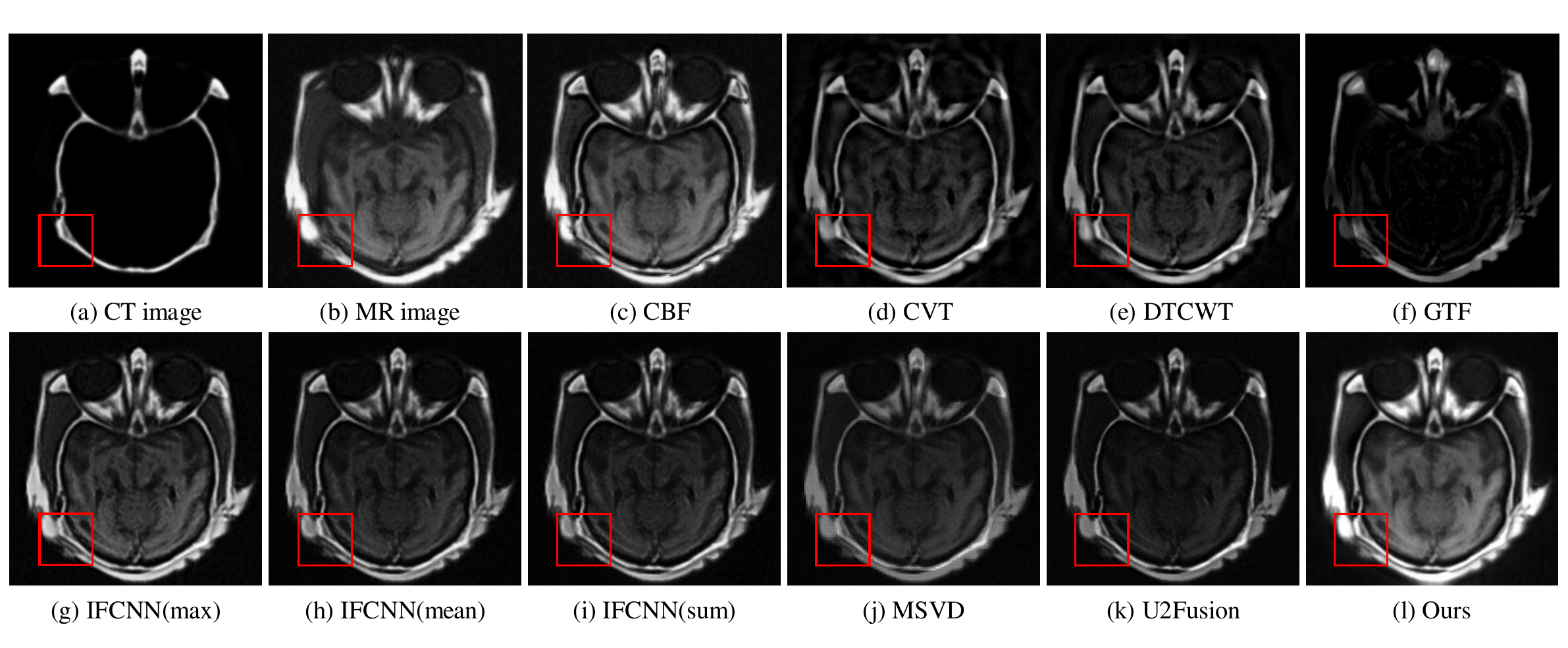}
		\caption{The medical image fusion experiment results of CT and MR images.}\label{fig-medical1}
	\end{center}
\end{figure*}
\begin{table*}[!t]
	\footnotesize
	\renewcommand{\arraystretch}{1.3}
	\caption{Quantitative evaluation results of medical image fusion tasks. The best three results are highlighted in {\color{red}{red}}, {\color{blue}{blue}} and {\color{brown}{brown}} fonts.}
	\label{eva-medical}
	\centering
	\begin{tabular}{cccccccccc}
		\hline
		Methods     & EI      & SF      & EN     & Q$^{ab/f}$   & SCD    & ~MS\_SSIM & MI      & SD      & VIF     \\
		\hline
		CBF         & \color{brown}{\textbf{82.1394}} & 22.5670 & 5.7775 & 0.5500 & 0.8316 & 0.8520    & 11.5550 & \color{brown}{\textbf{61.6619}} & 0.8147  \\
		CVT         & 79.0579 & 22.3262 & \color{blue}{\textbf{6.0096}} & 0.5229 & 0.7990 & 0.8859    & \color{blue}{\textbf{12.0191}} & 58.5920 & 0.8511  \\
		DTCWT       & 77.6919 & 22.4307 & \color{brown}{\textbf{5.8597}} & \color{brown}{\textbf{0.5559}} & 0.8036 & 0.8994    & \color{brown}{\textbf{11.7194}} & 58.8025 & \color{brown}{\textbf{0.8539}}  \\
		GTF         & 55.8531 & 16.3053 & 5.2286 & 0.4476 & 0.4973 & 0.7481    & 10.4572 & 50.7192 & 0.4657  \\
		MSVD        & 54.9092 & 19.9155 & 5.5083 & 0.4206 & 0.7683 & 0.8704    & 11.0166 & 53.8837 & 0.5566  \\
		IFCNN(max)  & \color{blue}{\textbf{83.1070}} & \color{blue}{\textbf{24.1949}} & 5.6613 & \color{blue}{\textbf{0.5737}} & \color{blue}{\textbf{1.1406}} & \color{red}{\textbf{0.9329}}    & 11.3226 & \color{blue}{\textbf{67.3829}} & \color{blue}{\textbf{0.9767}}  \\
		IFCNN(mean) & 76.0946 & 22.6930 & 5.5177 & 0.5466 & \color{brown}{\textbf{0.9077}} & \color{brown}{\textbf{0.9197}}    & 11.0354 & 58.2606 & 0.8062  \\
		IFCNN(sum)  & 76.8979 & \color{brown}{\textbf{22.9619}} & 5.5036 & 0.5416 & 0.8822 & 0.9165    & 11.0073 & 57.8363 & 0.8070  \\
		U2Fusion    & 69.4348 & 20.1861 & 5.4758 & 0.4687 & 0.7430 & 0.8967    & 10.9516 & 54.0862 & 0.7678  \\
		Ours        & \color{red}{\textbf{88.2840}} & \color{red}{\textbf{25.7036}} & \color{red}{\textbf{6.0503}} & \color{red}{\textbf{0.6289}} & \color{red}{\textbf{1.4161}} & \color{blue}{\textbf{0.9231}}    & \color{red}{\textbf{12.1005}} & \color{red}{\textbf{77.8152}} & \color{red}{\textbf{1.1499}} \\
		\hline
	\end{tabular}
\end{table*}
\begin{figure}[!t]
	\begin{center}
		\includegraphics[width=1\linewidth]{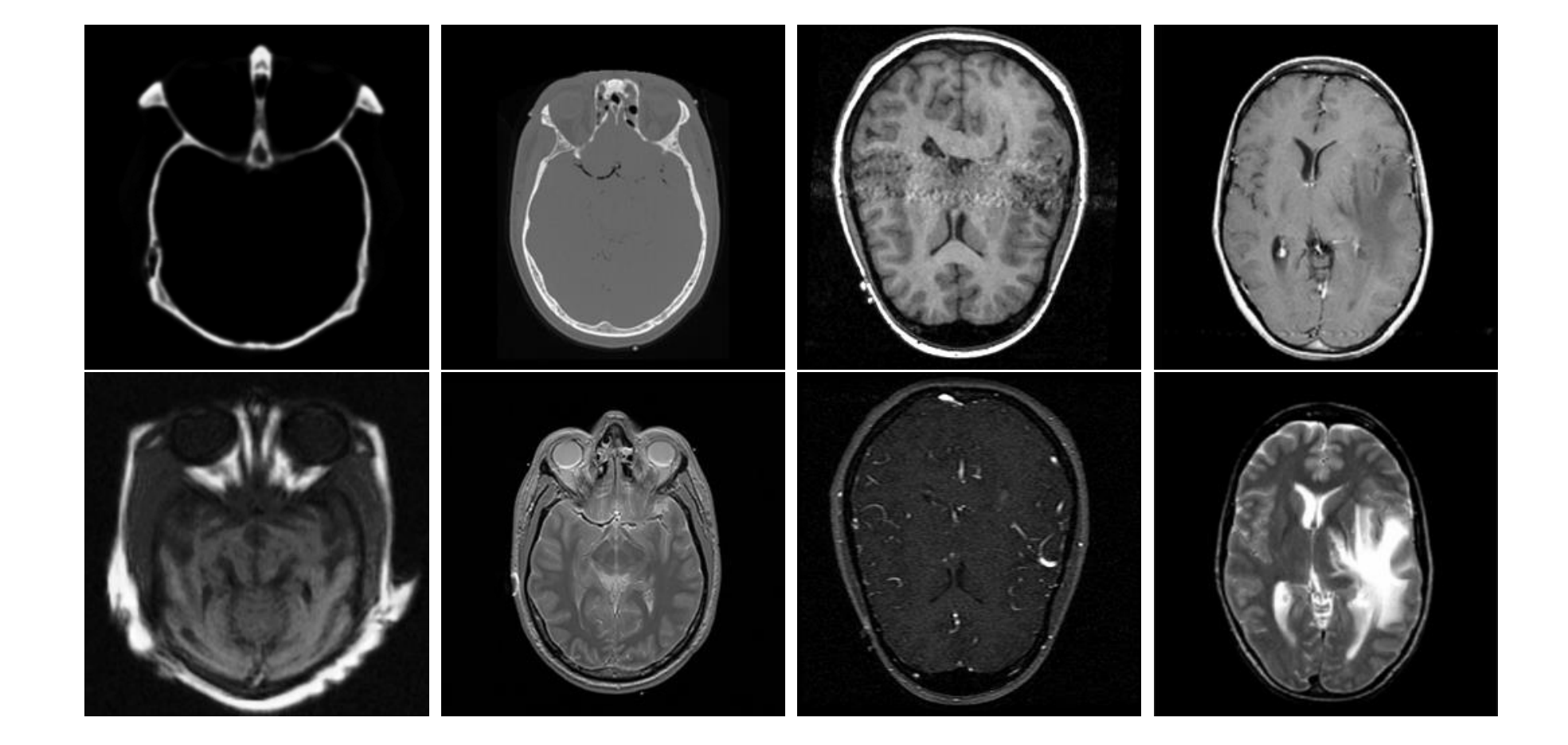}
		\caption{Four pairs of source images. The top row contains CT images, and the second row contains MR images.}\label{fig-medical}
	\end{center}
\end{figure}

In terms of objective evaluation, we compare with 12 methods, including classic and latest methods:Ratio of Low-pass Pyramid (RP) \cite{toet1989image}, Wavelet \cite{chipman1995wavelets}, Dual-Tree Complex Wavelet Transform (DTCWT) \cite{lewis2007pixel}, Curvelet Transform (CVT) \cite{nencini2007remote}, Multi-resolution Singular Value Decomposition (MSVD) \cite{naidu2011image}, gradient transfer and total variation minimization (GTF) \cite{ma2016infrared}, DenseFuse \cite{li2018densefuse}, DeepFuse \cite{prabhakar2017deepfuse}, a general end-to-end fusion network(IFCNN) \cite{zhang2020ifcnn}, NestFuse \cite{li2020nestfuse}, FusionGAN \cite{ma2019fusiongan}, U2Fusion \cite{xu2020u2fusion}.
In the quantitative comparison, we used the seven evaluation methods mentioned above.
All quantitative comparison results are shown in Table.\ref{eva-ir}. 
Our method achieved the best results on the four evaluation indicators (Q$^{ab/f}$, MS$_-$SSIM, SD, VIF), and achieved the second and third results on the remaining indicators (second: EI, CE; third: SF, EN).

\begin{table*}
	\footnotesize
	\renewcommand{\arraystretch}{1.3}
	\caption{The objective evaluation with different networks. The best results are highlighted in {\textbf{bold}} fonts.}
	\label{network}
	\centering
	\begin{tabular}{ccccccccc}
		\hline
		& EI      & CE     & SF     & EN     & Q$^{ab/f}$   & ~MS$_-$SSIM & SD      & VIF     \\
		\hline
		VIF-Net        & 34.7517 & \textbf{1.3231} & 8.2169 & 6.7120 & 0.4725 & \textbf{0.8568}    & 85.6736 & 0.5236  \\
		Ours-Net & \textbf{38.2261} & 1.3859 & \textbf{9.9340} & \textbf{6.6467} & \textbf{0.4939} & 0.8042    & \textbf{86.1420} & \textbf{0.6493} \\
		\hline
	\end{tabular}
\end{table*}
\begin{table*}[!t]
	\footnotesize
	\renewcommand{\arraystretch}{1.3}
	\caption{The objective evaluation with different loss functions. The best results are highlighted in {\textbf{bold}} fonts.}
	\label{loss}
	\centering
	\begin{tabular}{ccccccccc}
		\hline
		& EI      & CE     & SF      & EN     & Q$^{ab/f}$   & ~MS$_-$SSIM & SD      & VIF     \\
		\hline
		Ours(Mean)   & 38.2261 & 1.3859 & 9.9340  & 6.6467 & 0.4939 & 0.8042    & 86.1420 & 0.6493  \\
		Ours(Var) & \textbf{44.5137} & \textbf{1.1959} & \textbf{11.3889} & \textbf{6.8980} & \textbf{0.5436} & \textbf{0.9235}    & \textbf{89.8534} &\textbf{0.7556} \\
		\hline
	\end{tabular}
\end{table*}

\subsection{Multi-focus Image Fusion}\label{focus}
In Fig. \ref{fig-focus}, we present some examples of multi-focus images. 
The first row of image content is focused on the foreground, the second row of image content is focused on the background.
Therefore, the purpose of multi-focus image fusion is to fuse two images focused on different positions to obtain a fused image with clear content.
In the test, we used twenty pairs of multi-focus images to fuse.

We show the comparison of the visual fusion results of ten methods, including classic and latest methods: ASR \cite{liu2014simultaneous}, CBF \cite{kumar2015image}, CSR \cite{liu2016image}, DRPL \cite{li2020drpl}, RP$_-$SP \cite{liu2015general}, MGFF \cite{bavirisetti2019multi}, IFCNN \cite{zhang2020ifcnn} and U2Fusion \cite{xu2020u2fusion}. 
In terms of visual effects, various methods can achieve better fusion results. 
But our method is not trained on the multi-focus image data set, and it can achieve comparable results with other methods. 
This shows that our fusion module and network structure have been fully trained in a single data set to learn quite transferable feature extraction capabilities as well as the fusion rules.

\begin{figure}[!t]
	\begin{center}
		\includegraphics[width=1\linewidth]{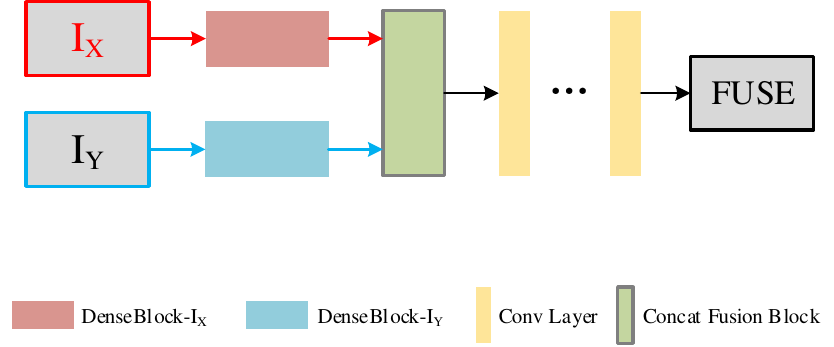}
		\caption{The framework of VIF-Net.}\label{vif}
	\end{center}
\end{figure}

In the quantitative comparison, we use the seven evaluation methods mentioned above.
All quantitative comparison results are shown in Table \ref{eva-focus}.
Our method achieved the best results on the five evaluation indicators (EI, SF, SCD, SD, VIF), and achieved the second results on the remaining indicators (EN, MI).

\subsection{Medical Image Fusion}\label{medical}
In Fig. \ref{fig-medical}, We visualize some medical images that need to be fused, including CT images and MR images. 
The purpose of medical image fusion is similar to the task of infrared and visible image fusion. 
It combines the salient information in the two source images to generate a fused image to assist visual judgment. 
This has a significant effect on improving the results of medical diagnosis. 
In the test, we use eight pairs of medical images (CT and MR) to fuse.

In Fig. \ref{fig-medical1}, we show the fusion results of CT and MR images obtained by ten methods, including traditional methods and deep learning-based methods: CBF \cite{kumar2015image}, CVT \cite{nencini2007remote}, DTCWT \cite{lewis2007pixel}, GTF \cite{ma2016infrared}, IFCNN \cite{zhang2020ifcnn}, MSVD \cite{naidu2011image}, U2Fusion \cite{xu2020u2fusion} and ours. 
Different features are shown in the red boxes of the two source images. 
We hope to obtain these two different kinds of information at the same time in the fused image. 
Among the fusion results of ten methods, some methods did not preserve both information well (f, j). 
Others use fusion to reduce the saliency of the information in the original image. 
Our method guarantees the salient information of the two source images and achieves a better fusion effect.

In Table \ref{eva-medical}, we use nine objective evaluation methods to compare the fusion results. 
The red font ratio indicates the best result. 
Our method performs the best in eight indicators (EI, SF, EN, Q$^{ab/f}$, SCD, MI, SD, VIF) and ranks second in the remaining one metric (~MS\_SSIM).
\subsection{Ablation Study}\label{ablation}
\subsubsection{Network structure}
In order to verify the effectiveness of our network structure, the VIF-Net \cite{hou2020vif} is used to compare to our method, which uses a similar loss function and training method. 
The framework of VIF-Net is shown in Fig \ref{vif}.  
In order to compare with VIF-Net, we keep the loss function consistent with it, so that our method is only different from VIF-Net in the network structure. 
At the same time, we chose the same cascade fusion for the fusion module.
As shown in Table \ref{network}, our method has achieved advantages in multiple objective evaluations, which also shows the effectiveness of the network structure. 
The network structure we proposed obtains the multi-scale information of the image through scale transformation.
Besides, multi-scale information is enriched through feature aggregation, so better performance is obtained.
\subsubsection{Loss function}
In terms of verifying the effect of the network structure, we use the same loss function as VIF-Net which uses the average of the pixel values of each sliding window as the criterion for selecting the patch of source image. 
Then we improve the loss function, using variance as the selection criterion for each sliding window.
While keeping the network structure and fusion module unchanged, we compare the impact of the results produced by different loss functions.
As Table \ref{loss} shows, the result of our proposed loss function is ahead of the original loss in all aspects.
Compared with the mean, variance can characterize the distribution of information, which makes the model have better performance and a wider application range.

\begin{table*}[!t]
	\footnotesize
	\renewcommand{\arraystretch}{1.3}
	\caption{The objective evaluation with different fusion rules. The best results are highlighted in {\textbf{bold}} fonts.}
	\label{fusion}
	\centering
	\begin{tabular}{ccccccccc}
		\hline
		& EI      & CE     & SF      & EN     & Q$^{ab/f}$   & ~MS$_-$SSIM & SD      & VIF     \\
		\hline
		Add  & 42.7438 & \textbf{0.9699} & 11.0250 & 6.7928 & 0.4539 & 0.9165 & 91.2557 & 0.6893 \\
		Concat            & 44.5137 & 1.1959 & 11.3889 & \textbf{6.8980} & 0.5436 & \textbf{0.9235}    & 89.8534 & 0.7556  \\
		Mapping & \textbf{45.5980} & 1.2941 & \textbf{11.5680} & 6.8435 & \textbf{0.5581} & 0.9164    & \textbf{93.0976} & \textbf{0.7897}  \\
		\hline
	\end{tabular}
\end{table*}
\begin{table*}[!t]
	\footnotesize
	\renewcommand{\arraystretch}{1.3}
	\caption{The objective evaluation with different network depth. The best results are highlighted in {\textbf{bold}} fonts.}
	\label{layer}
	\centering
	\begin{tabular}{ccccccccc}
		\hline
		& EI      & CE     & SF      & EN     & Q$^{ab/f}$   & ~MS$_-$SSIM & SD      & VIF     \\
		\hline
		3-layers &\textbf{ 45.5980} & 1.2941 & \textbf{11.5680} & 6.8435 & 0.5581 & 0.9164    & \textbf{93.0976} & \textbf{0.7897}  \\
		4-layers & 44.0749 & \textbf{1.2546} & 11.2483 & \textbf{6.9105} & \textbf{0.5604} & \textbf{0.9242}    & 88.8935 & 0.7623  \\
		5-layers & 44.2599 & 1.2596 & 11.2986 & 6.8597 & 0.5408 & 0.9022    & 91.6852 & 0.7345  \\
		\hline
	\end{tabular}
\end{table*}
\subsubsection{Fusion method}
In addition to the network structure and loss function, the design of the fusion method is another core of the image fusion algorithm. 
After adopting our network structure and loss function, we compared three methods, including direct addition, direct cascade, and dense feature mutual mapping. 
The results are shown in Table \ref{fusion}. 
The best fusion image performance can be obtained by using the feature mutual mapping module we proposed for fusion.
As the simplest method, direct addition has achieved good results in some algorithms. 
However, it may cause the problem of neutralization of the pixel value of the corresponding position, which makes the poor quality of the fused image. 
The concatenation fusion method allows the model to actively learn the fusion weights during the training process, but the concatenation is too simple to learn the fusion method we need well.
The feature mutual mapping module we proposed learns a satisfactory fusion method in training through more complex structures, and can be directly transferred to more fusion tasks.
\subsubsection{Network depth}
After determining that our network structure and loss function are effective, we continue to explore the impact of the depth of the network on the fusion performance.
As shown in Fig.\ref{main-fram}, our encoder network is composed of a Conv layer and two res-blocks which we define its depth as three layers.  
The initial network depth is determined by our fusion method, so we try to prove whether a deeper network can deliver better results.

Every time we deepen the network, we add downsampling and res at the encoder, and the decoder part will also increase the corresponding aggregation structure. 
In Table \ref{layer}, we show the results of two experiments after deepening the network, which are 4-layers and 5-layers. 
The results show that deepening the network does not significantly boost fusion performance.
This also shows that after the network depth reaches a certain level, continuing to deepen the network will not have a positive effect on the performance improvement, and may even cause a negative effect.
\section{Conclusion}\label{conclusion}
We proposed a novel image fusion method based on feature mutual mapping and multi-scale autoencoder.
In the feature mutual mapping and fusion module, the network can adaptively learn the fusion strategy during the training process by finding the mapping relationship between the features of the two source images.
The multi-scale autoencoder and the aggregation operation of different scale features in the decoding process make the fusion fully performed at different granularities.
Experiments on a variety of tasks and their public data sets prove the effectiveness and versatility of our method.



\ifCLASSOPTIONcaptionsoff
  \newpage
\fi



\bibliographystyle{IEEEtran}
\bibliography{ref.bib}

\end{document}